\documentclass{article}

\PassOptionsToPackage{numbers, compress}{natbib}
\usepackage[preprint]{neurips_2026}

\usepackage[utf8]{inputenc} 
\usepackage[T1]{fontenc}    
\usepackage{hyperref}       
\usepackage{url}            
\usepackage{booktabs}       
\usepackage{amsfonts}       
\usepackage{nicefrac}       
\usepackage{microtype}      
\usepackage{xcolor}         
\usepackage{amsmath}
\usepackage{enumitem}
\usepackage{wrapfig}    
\usepackage{microtype}
\usepackage{graphicx}
\usepackage{subcaption}
\usepackage{tabularx} 
\usepackage{multirow} 
\usepackage[table]{xcolor}
\usepackage{listings}
\usepackage{fvextra}

\title{Reasoning Before Diagnosis: Physician-Inspired Structured Thinking for ECG Classification}

\author{%
Yang Wu$^{1}$\thanks{\,Equal contribution.},~~~
Xiaoyan Yuan$^{2,3}$\footnotemark[1],~~~
\textbf{Hau-San Wong}$^{1}$\footnotemark[2],~~~
\textbf{Xiping Hu}$^{2,3}$\thanks{\,Corresponding authors.} \\[0.75em]
$^{1}$City University of Hong Kong ~~~
$^{2}$Beijing Institute of Technology\\[0.3em]
$^{3}$Shenzhen MSU-BIT University
}

\begin{document}

\maketitle

\begin{abstract}
   Electrocardiogram (ECG) diagnosis in clinical practice relies on structured reasoning over multiple hierarchical aspects, including cardiac rhythm, conduction properties, waveform morphology, and overall diagnostic impression. However, most existing approaches predict labels directly from ECG signals without explicit clinical reasoning, resulting in opaque decisions that lack clinical alignment. To bridge this gap, we propose \textbf{CardioThink}, a physician-inspired multimodal large language model (MLLM) framework that explicitly models the diagnostic reasoning process through human-interpretable intermediate stages (rhythm, conduction, morphology, and impression) to derive final classification results. Furthermore, we introduce \textbf{S}tructured \textbf{S}et \textbf{P}olicy \textbf{O}ptimization (\textbf{SSPO}) to jointly optimize adherence to this structured reasoning format and the accuracy of variable-size diagnostic sets, without requiring manually annotated reasoning traces. Extensive experiments on diverse ECG benchmarks demonstrate the significant superiority of our approach in diagnostic accuracy, while simultaneously providing interpretable clinical reasoning. Notably, reasoning quality evaluations confirm that SSPO substantially enhances the clinical validity of the generated rationales. These findings reveal that moving beyond direct label prediction toward structured reasoning offers a more clinically aligned direction for future ECG modeling.
\end{abstract}

\section{Introduction}

Electrocardiogram (ECG) analysis is a diagnostic cornerstone, routinely used to identify diverse cardiac conditions~\cite{naturereview, NC}. In clinical practice, this is inherently a reasoning-driven process: clinicians do not make single-step decisions but progressively synthesize cardiac rhythm, conduction properties, and waveform morphology~\cite{naturemedicine,Brushe064389}. This step-by-step reasoning process is crucial for aligning automated analysis with the essence of human diagnosis.

However, a fundamental limitation persists across existing ECG analysis paradigms: they are largely designed to derive diagnostic categories directly from inputs without explicit intermediate inference.
Standard discriminative models, such as CNNs and Transformers~\cite{yuan2025enhancing,yuan2025reading,MVMNet,MRMNet}, as well as recent self-supervised representation learning frameworks~\cite{DBEAT,ECG2TOK,MERL,MELP}, have demonstrated strong efficacy by learning high-dimensional mappings from signals to labels. Yet, these approaches inherently neglect the causal structure of clinical diagnosis. Rather than encoding medical logic involving the integration of specific morphological and rhythmic cues, these models tend to exploit spurious statistical correlations~\cite{li2025generative}. This decoupling imposes a critical constraint: the lack of explicit reasoning is not merely a transparency deficit but a fundamental bottleneck that limits the achievable accuracy. Consequently, in challenging regimes requiring rigorous clinical deduction, the reliance on purely statistical patterns severely limits model reliability.

Recent progress in Multimodal Large Language Models (MLLMs) has enabled the generation of automated ECG reports~\cite{MEIT,ECGCHAT,PULS,GEM,QHEART,jin2026ecg}. However, despite this generative capability, their decision-making mechanisms remain inherently opaque. Synthesizing natural language descriptions does not equate to structured clinical reasoning. Internally, intermediate clinical concepts remain entangled within latent representations rather than acting as distinct, controllable variables for inference. Externally, the generated reports serve merely as unstructured parallel outputs, failing to provide explicit structural support for the diagnostic predictions. Consequently, these frameworks still operate as black boxes, prioritizing predictive performance over interpretable clinical logic.

In this work, we argue that effective automated ECG diagnosis should move beyond direct label prediction and instead explicitly incorporate clinical reasoning into the modeling process.
To this end, we introduce CardioThink, a physician-inspired reasoning framework for ECG classification. It decomposes diagnosis into structured reasoning stages—rhythm analysis, conduction assessment, and waveform morphology—synthesizing these insights to derive the final classification.

Furthermore, to alleviate the annotation burden of reasoning trajectories, we introduce Structured Set Policy Optimization (SSPO), which enhances the quality of structured reasoning and optimizes variable-size diagnostic outputs without requiring supervision for intermediate reasoning steps.
Experiments across diverse ECG classification tasks show that CardioThink consistently outperforms prior methods and produces interpretable intermediate thinking, suggesting explicit reasoning as a more effective and clinically aligned modeling paradigm. Our work makes three primary contributions:

\begin{itemize}
  
  \item We propose CardioThink, a physician-inspired ECG classification framework that explicitly models clinical reasoning, together with Structured Set Policy Optimization (SSPO) to enhance reasoning quality and diagnostic accuracy.
  
  \item We construct a reasoning-oriented ECG dataset featuring structured diagnostic reasoning paths, providing explicit supervision signals to facilitate interpretable clinical reasoning.
  
  \item We demonstrate consistent performance improvements across large-scale ECG benchmarks (up to +22.73\% F1) and validate the clinical validity of the generated reasoning, highlighting CardioThink’s potential for reliable and interpretable clinical decision support.
\end{itemize}

\section{Related Work}
\subsection{Representation Learning for ECG Analysis}
Recent progress in automated ECG analysis has evolved from supervised baselines to advanced representation learning paradigms. Early deep learning approaches relied on end-to-end supervised training to map raw signals directly to diagnostic labels using convolutional or transformer-based architectures~\cite{MVMNet,yuan2025enhancing,MRMNet,yuan2025reading}. To overcome generalization limits, self-supervised strategies subsequently emerged, treating ECG as a cardiac language and employing objectives like masked token reconstruction to capture high-level structural patterns independent of labels~\cite{ECG2TOK,HeartLang,tracing}. More recently, multimodal frameworks align ECG signals with paired clinical reports via contrastive or matching objectives, integrating semantic medical knowledge to support zero-shot capabilities~\cite{MERL,DBEAT,MELP,SuPreME}. 

Despite these methodological variations, a fundamental limitation persists, as these approaches universally rely on a discriminative modeling paradigm where diagnostic decisions are derived directly from learned embeddings. This framework effectively bypasses explicit clinical reasoning, replacing it with opaque statistical correlations between raw signals and labels~\cite{zhou2025robustness}. Consequently, these systems operate largely as black boxes, prioritizing predictive performance over the structured, transparent deduction essential for clinical reliability.

\subsection{MLLMs for ECG Analysis}

Recently, Multimodal Large Language Models (MLLMs) have been increasingly explored for ECG analysis, primarily to enable ECG-conditioned report generation, retrieval, and interactive interpretation~\cite{MEIT, QHEART,HE2025102963, yang2025ecg, jin2025uniecg}. Some studies focus on aligning ECG signals with clinical text to support multimodal understanding. For instance, ECG-Chat~\cite{ECGCHAT} employs contrastive learning to align ECG signals with clinical reports, enabling report generation, conversational interaction, and zero-shot retrieval. Other works extend this paradigm to visual ECG representations. PULSE~\cite{PULS} introduces a large-scale ECG image instruction-tuning dataset and trains an ECG-specialized multimodal language model, demonstrating strong performance on diverse ECG image interpretation tasks. Building on these efforts, GEM~\cite{GEM} further unifies ECG signals, ECG images, and text within a multimodal framework, and evaluates grounded ECG understanding by linking diagnostic statements to fine-grained waveform evidence. Furthermore, ECG-R1~\cite{jin2026ecg} integrates reinforcement learning to mitigate hallucinations in open-ended narrative clinical reports, utilizing DAPO~\cite{yu2025dapo} and evidence-coverage rewards based on step-by-step evidence verification.

\begin{figure*}[t!]
    \centering
    \includegraphics[width=1\linewidth]{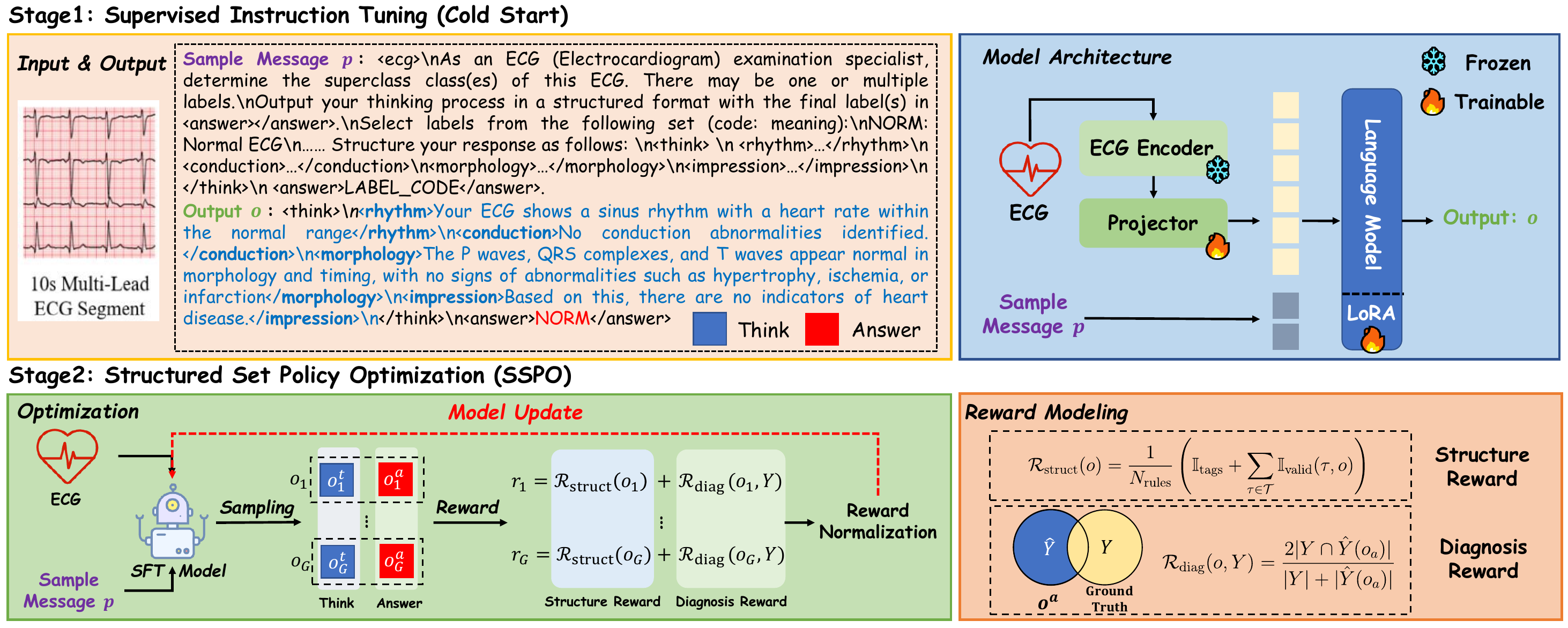}
    \caption{Overall pipeline of CardioThink. Our model is trained in two stages: supervised instruction tuning (Cold Start) and reinforcement learning (SSPO) on ECG data. In the cold-start stage, the model is initialized to generate structured ECG reasoning traces along with diagnostic predictions. In the reinforcement learning stage, training is guided by two reward signals: a structure reward that enforces valid reasoning format, and a diagnosis reward that evaluates diagnostic correctness.}
    \label{fig:outline}
\end{figure*}
Despite these advances, existing MLLM-based approaches typically treat ECG classification and report generation as independent tasks. While classification is performed via direct label prediction on separate datasets, report generation is mainly used to demonstrate interpretability or clinical plausibility. Consequently, the generated text functions as a post-hoc explanation rather than actively participating in the production of diagnostic labels. Because of this, diagnostic reasoning remains implicit in latent representations instead of explicitly driving the decision process, leaving a gap between model predictions and the structured reasoning required in clinical ECG interpretation.

\section{Methods}

\subsection{Problem Definition}

Formally, let $\mathbf{X} \in \mathbb{R}^{T \times C}$ denote a multivariate ECG recording with length $T$ and $C$ leads. Unlike discriminative paradigms that map $\mathbf{X}$ to a fixed label set, we formulate ECG analysis as a \textit{conditional sequence generation} task, approximating the target distribution $p_\theta(\mathbf{Y} \mid \mathbf{X})$ with an auto-regressive model, where $\mathbf{Y} = (y_1, \dots, y_K)$ is a structured token sequence. Here, $\mathbf{Y}$ encodes both the intermediate reasoning process (physician-like rationale) and the final diagnosis. Assuming target sequences are conditionally independent across samples in a training dataset $\mathcal{D}$, the model is trained by maximizing the expected conditional log-likelihood objective: $\mathbb{E}_{(\mathbf{X}, \mathbf{Y}) \sim \mathcal{D}} \left[ \sum_{i=1}^{K} \log p_\theta(y_i \mid y_{<i}, \mathbf{X}) \right].$
This formulation natively captures the dependencies between underlying ECG patterns, sequential clinical reasoning, and the ultimate diagnostic decisions.

\subsection{Reasoning-Oriented Data Construction}
\label{sec:data_construction}
\begin{wrapfigure}{r}{0.5\textwidth} 
    \vspace{-10pt} 
    \centering
    \includegraphics[width=\linewidth]{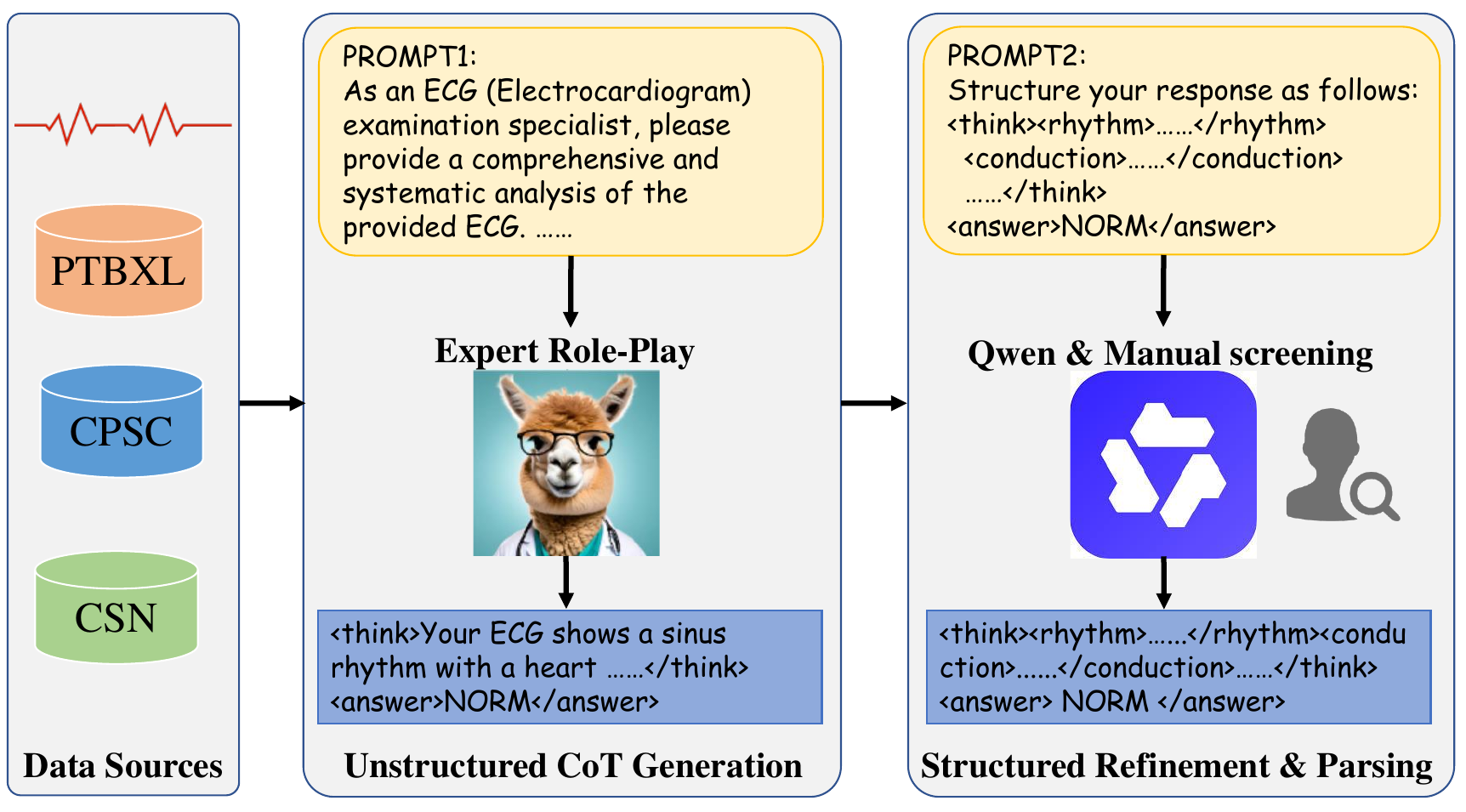}
    \caption{The reasoning-oriented ECG data pipeline: (1) generating unstructured analyses via expert role-play from multi-source ECGs; (2) structured parsing to extract diagnostic reasoning paths.}
    \label{fig:data_construction}
    \vspace{-13pt} 
\end{wrapfigure}

To endow the model with ECG reasoning capabilities, we constructed a dataset of high-quality, structured diagnostic reasoning paths using a scalable, model-guided approach. As illustrated in Figure~\ref{fig:data_construction}, the pipeline consists of two stages. First, following previous works~\cite{dai2026qoqmed, liu2025fleming}, we leveraged the PTB-XL, CPSC, and CSN datasets and employed ``Expert Role-Playing'' prompts to guide ECG-Chat-13B~\cite{ECGCHAT} in simulating cardiologist diagnostics. 
This process yielded a comprehensive collection of ECG analyses. To ensure data reliability at scale, we developed a semi-automated cleaning pipeline informed by the manual inspection of a data subset. This refinement focused on two primary aspects: (1) \textbf{Structural Integrity}: We employed Qwen-Plus~\cite{yang2025qwen3} with strictly constrained prompts to reformat the raw ECG analysis outputs into a standardized four-tier Chain-of-Thought (CoT) framework (Rhythm, Conduction, Morphology, and Impression). To prevent hallucinations, the introduction of any new information was explicitly prohibited. (2) \textbf{Natural Language Refinement}: Automated scripts were deployed to standardize ambiguous placeholders (e.g., mapping “<conduction>None</conduction>” to “<conduction>No conduction abnormalities identified.</conduction>") and to filter out samples lacking clinical impressions. This rigorous curation process ultimately yielded a highly refined dataset for model initialization.

\subsection{Model Architecture}
Figure~\ref{fig:outline} presents the model architecture of CardioThink, a MLLM framework designed for interpretable ECG classification. Built upon LLaVA-7B~\cite{liu2024improved}, it integrates a specialized physiological encoder and a modality alignment projector. This design empowers the MLLM to perceive high-dimensional ECG signals, generating explicit diagnostic reasoning alongside classification decisions.

\noindent\textbf{ECG Encoder.} 
To derive robust pathological representations from high-dimensional physiological signals, we employ a 1D-Vision Transformer architecture. Given an input ECG series $\mathbf{X}$, we first tokenize the signal into a sequence of non-overlapping patches. Following~\cite{ECGCHAT}, we leverage a pre-trained encoder $\Phi_{\text{enc}}$ comprising 12 Transformer layers with a hidden dimension $d_{\text{enc}}=768$. This architecture allows the model to capture long-range temporal dependencies within cardiac cycles. Formally, we obtain the patch-level feature representations $\mathbf{Z}_{\text{ecg}}$ as:
\begin{equation}
    \mathbf{Z}_{\text{ecg}} = \Phi_{\text{enc}}(\mathbf{X}) \in \mathbb{R}^{L \times d_{\text{enc}}},
\end{equation}
where $L$ denotes the sequence length. This ensures the encoder possesses a rich semantic understanding of cardiac pathologies prior to instruction tuning.

\noindent\textbf{Modality Alignment.} 
A critical challenge in Multimodal LLMs is bridging the semantic gap between continuous signal embeddings and the discrete token space of the language model. Following~\cite{GEM}, we introduce a projection module $f_{\text{proj}}$, instantiated as a two-layer multilayer perceptron with GELU activation~\cite{hendrycks2016gaussian}. This projector maps encoded ECG features into the word embedding space:
\begin{equation}
    \mathbf{H}_{\text{ecg}} = f_{\text{proj}}(\mathbf{Z}_{\text{ecg}}) \in \mathbb{R}^{L \times d_{\text{llm}}},
\end{equation}
where $d_{\text{llm}}$ denotes the hidden dimension of the LLM backbone. The resulting $\mathbf{H}_{\text{ecg}}$ acts as a sequence of continuous embeddings compatible with the LLM's input space, enabling the model to process physiological signals jointly with textual tokens.

\subsection{Supervised Instruction Tuning (Cold Start)}
Prior to reinforcement learning, we conduct Supervised Fine-Tuning on the cold start dataset to equip the model with initial diagnostic capabilities and structured reasoning behaviors. 

\noindent\textbf{Input Formulation.} 
For each ECG sample $\mathbf{X}$, let $q$ denote the user query. The tokenized query embeddings are denoted as $\mathbf{H}_{\text{query}}$. To fuse the modalities, we concatenate the projected ECG features $\mathbf{H}_{\text{ecg}}$ with the query embeddings to form the unified input sequence $\mathbf{I}$:
\begin{equation}
    \mathbf{I} = [\mathbf{H}_{\text{ecg}}, \mathbf{H}_{\text{query}}],
    \label{eq:sft_input}
\end{equation}
where $[\cdot, \cdot]$ denotes the concatenation operation along the sequence dimension.

\noindent\textbf{Optimization.} 
The target output $\mathbf{Y} = (y_1, \dots, y_K)$ is constructed to strictly adhere to the $\texttt{<think>}$ and $\texttt{<answer>}$ format, where the \texttt{<think>} block encapsulates structured clinical reasoning. This design explicitly separates the intermediate rationale from the final diagnosis. We fine-tune the trainable LoRA~\cite{hu2022lora} parameters $\theta$ by minimizing the standard negative log-likelihood objective:
\begin{equation}
    \mathcal{L}_{\text{SFT}}(\theta) = - \sum_{t=1}^{M} \log P_\theta(y_t \mid \mathbf{I}, y_{<t}),
\end{equation}
where $P_\theta$ is the probability distribution parameterized by the LLM. This process initializes the policy $\pi_{\text{sft}}$ for the subsequent reinforcement learning stage.

\subsection{Structured Set Policy Optimization (SSPO)}
Following SFT, we propose SSPO to further enhance the model's reasoning capabilities through reinforcement learning. Building on Group Relative Policy Optimization (GRPO)~\cite{shao2024deepseekmath}, an efficient alternative to Proximal Policy Optimization (PPO)~\cite{schulman2017proximal}, SSPO employs a composite reward function to bolster structural rigor and diagnostic accuracy.

\noindent\textbf{Reward Formulation.}
The policy $\pi_{\theta}$ generates a group of $G$ outputs $\{o_i\}_{i=1}^{G}$ for a unified input $\mathbf{I}$ (Eq. \ref{eq:sft_input}). For each output $o_i$, we calculate a scalar reward $r_i$ defined as:
\begin{equation}
    r_i = \mathcal{R}_{\text{struct}}(o_i) + \mathcal{R}_{\text{diag}}(o_i, Y).
\end{equation}
The specific design of these two components is as follows:

\begin{itemize}[leftmargin=*, labelsep=0.5em]
    \item \textbf{Structure Reward ($\mathcal{R}_{\text{struct}}$):} To strictly enforce the chain-of-thought format, we require the presence of correct delimiters (e.g., \texttt{<think>}, \texttt{<answer>}) and four mandatory clinical sections $\mathcal{T}$: \texttt{rhythm}, \texttt{conduction}, \texttt{morphology}, and \texttt{impression}. The reward represents the normalized completion rate of these $N_{\text{rules}}=5$ constraints:
    \begin{equation}
        \mathcal{R}_{\text{struct}}(o) = \frac{1}{N_{\text{rules}}} \left( \mathbb{I}_{\text{tags}} + \sum_{\tau \in \mathcal{T}} \mathbb{I}_{\text{valid}}(\tau, o) \right),
    \end{equation}
    where $\mathbb{I}_{\text{tags}}$ validates the tag hierarchy, and $\mathbb{I}_{\text{valid}}(\tau, o)$ denotes a non-empty section $\tau$.
    \item \textbf{Diagnosis Reward ($\mathcal{R}_{\text{diag}}$):} We formulate ECG diagnosis as a variable-size set prediction task. We decompose the model output $o$ into a reasoning trace $o_t$ (corresponding to the \texttt{<think>} section) and a final prediction $o_a$ (corresponding to the \texttt{<answer>} section). Let $Y$ be the ground-truth set and $\hat{Y}(o_a)$ denote the diagnostic set parsed specifically from $o_a$. To evaluate the agreement between these sets, we employ the Dice coefficient as the reward signal:
        \begin{equation}
            \mathcal{R}_{\text{diag}}(o, Y) = \frac{2 |Y \cap \hat{Y}(o_a)|}{|Y| + |\hat{Y}(o_a)|},
        \end{equation}
        where $|\cdot|$ denotes the set cardinality. This formulation treats all diagnostic categories with equal importance and penalizes both missed diagnoses and hallucinations through the denominator.
\end{itemize}

\paragraph{Optimization Objective.}
We employ GRPO for policy updates. For the generated group $\{o_i\}_{i=1}^{G}$, we normalize rewards via group statistics to compute relative advantage and reduce variance:
\begin{equation}
    A_i = \frac{r_i - \mu_r}{\sqrt{\sigma_r^2 + \epsilon}},
\end{equation}
where $\mu_r$ and $\sigma_r^2$ are the mean and variance of $\{r_1, \dots, r_G\}$. The policy is updated by maximizing the following objective over the generated group:
\begin{equation}
\label{eq:sspo_loss}
\begin{aligned}
    \mathcal{L}_{\text{GRPO}}(\theta) = \frac{1}{G} \sum_{i=1}^{G} \Big[ & \min \left( \rho_i A_i, \text{clip}(\rho_i, 1-\epsilon_c, 1+\epsilon_c) A_i \right) - \beta D_{\text{KL}}(\pi_{\theta}(o_i|\mathbf{I}) \| \pi_{\text{ref}}(o_i|\mathbf{I})) \Big],
\end{aligned}
\end{equation}
where $\rho_i = \frac{\pi_{\theta}(o_i | \mathbf{I})}{\pi_{\theta_{\text{old}}}(o_i | \mathbf{I})}$ is the likelihood ratio, and $D_{\text{KL}}$ ensures the model retains linguistic fluency by penalizing deviation from the SFT model $\pi_{\text{ref}}$.

\section{Experiments}
\subsection{Experimental Datasets}
To comprehensively evaluate the effectiveness of our proposed method, we conduct extensive experiments across three diverse ECG datasets that collectively encompass six distinct clinical tasks. Specifically, \textbf{PTB-XL}~\cite{wagner2020ptb} is a large-scale dataset comprising 21,837 10-second 12-lead recordings from 18,885 patients sampled at 500 Hz, supporting four multi-label classification tasks: Superclass (5 categories), Subclass (23 categories), Form (19 categories), and Rhythm (12 categories). Furthermore, \textbf{CPSC2018}~\cite{liu2018open} consists of 6,877 12-lead recordings sampled at 500 Hz, with durations ranging from 6 to 60 seconds across nine diagnostic labels. Finally, the \textbf{Chapman--Shaoxing--Ningbo (CSN)}~\cite{zheng2020optimal,zheng2022large} dataset includes 10-second 12-lead recordings sampled at 500 Hz, yielding 23,026 valid recordings across 38 labels after filtering out unknown annotations.  We adopt the official splits for all datasets to ensure fair comparisons.

\subsection{Baseline Methods}
To validate the effectiveness of our proposed approach, we conduct a comprehensive comparison against two categories of multimodal ECG methods:
(1) Pretrained discriminative models, which learn representations from paired ECG-report data, such as MERL~\cite{MERL} and MELP~\cite{MELP}, and are subsequently fine-tuned on downstream ECG classification tasks; and
(2) Multimodal generative ECG models, including QoQ-Med~\cite{dai2026qoqmed}, ECG-Chat~\cite{ECGCHAT}, PULSE~\cite{PULS}, and GEM~\cite{GEM}, which perform ECG interpretation via language modeling by mapping multimodal ECG inputs to diagnostic text, from which classification decisions are derived. 

\subsection{Experimental Settings}
\label{sexseetting}
The cold-start phase is conducted with a batch size of 16 for 4 epochs for all datasets, using a learning rate of $2 \times 10^{-4}$. For the reinforcement learning phase, we set a learning rate of $2 \times 10^{-5}$ over 8 epochs with a batch size of 1 and 4 gradient accumulation steps, generating 4 candidate responses per query. All models are trained on two NVIDIA A800 GPUs. Following ECG-Chat~\cite{ECGCHAT}, we adopt Precision, Recall, and F1-score as metrics.

\begin{table*}[t!]
\centering
\caption{Performance comparison on six ECG classification benchmarks.
Best results are shown in \textbf{bold} and second-best results are \underline{underlined}. "Gain" refers to the performance improvement.}
\setlength{\tabcolsep}{2pt} 

\resizebox{\textwidth}{!}{
\begin{tabular}{ l *{18}{c} }
\toprule
\multirow{2}{*}{Model} 
& \multicolumn{3}{c}{\textbf{PTBXL-Super}}
& \multicolumn{3}{c}{\textbf{PTBXL-Sub}}
& \multicolumn{3}{c}{\textbf{PTBXL-Form}}
& \multicolumn{3}{c}{\textbf{PTBXL-Rhythm}}
& \multicolumn{3}{c}{\textbf{CPSC2018}}
& \multicolumn{3}{c}{\textbf{CSN}} \\
\cmidrule(lr){2-4} \cmidrule(lr){5-7} \cmidrule(lr){8-10} \cmidrule(lr){11-13} \cmidrule(lr){14-16} \cmidrule(lr){17-19}

& Prec. & Rec. & F1
& Prec. & Rec. & F1
& Prec. & Rec. & F1
& Prec. & Rec. & F1 
& Prec. & Rec. & F1
& Prec. & Rec. & F1\\
\midrule

\rowcolor{gray!15}
\multicolumn{19}{c}{\textcolor{gray}{\textit{Pretrained Discriminative Models}}} \\
MERL~\cite{MERL} &68.21 &\textbf{75.30} &	69.89 &	37.21 &	50.16 &	43.04 &	27.05 &\underline{45.07} &	32.91 &	47.15 &	47.15 &	51.78 &	65.86 &	64.31 &	63.86 &	36.51 &	44.35 &	41.43 \\
MELP~\cite{MELP} & 56.19 &72.26 &	60.60 &	28.57 &	42.19 &	33.19 &	15.08 &	44.52 &	20.97 &	21.29 &	43.65 &	21.30 &	27.44 &\underline{65.04} &	33.85 &	19.14 &	31.37 &	24.10 \\
\hline
\rowcolor{gray!15}
\multicolumn{19}{c}{\textcolor{gray}{\textit{Generative Models based on MLLM}}} \\
QoQ-Med~\cite{dai2026qoqmed} & 44.70 & 34.60 & 39.01 & 43.14 & 30.24 & 35.59 & 19.68 & 15.56 & 17.38 & 26.66 & 27.66 & 27.15 & 10.58 & 9.22 & 9.86 & 13.36 & 12.64 & 12.99 \\
GEM~\cite{GEM} & 41.15 &	36.71 &	38.80 &	46.06 &	36.22 &	40.55 &	\textbf{64.44} &	9.05 &	15.88 &	83.29 &	83.29 &	83.29 &	30.40 &	30.43 &	30.41 &	26.78 &	28.38 &	27.56 \\
PLUSE~\cite{PULS} &63.01	&67.12	&65.00	&60.16	&38.37	&46.85	&37.19	&18.73	&24.91	&73.60	&74.54	&74.07	&39.33	&28.75	&33.22 &43.37	&34.32	&38.32 \\
ECG-Chat~\cite{ECGCHAT} & 43.47 &	49.21 &	46.16&	35.09 & 36.91 &	35.98 & 25.54&	15.11&	17.35&	54.76&	40.02&	43.39 &12.32	&35.65	&18.31&10.30&	46.88	&16.89 \\
\textbf{CardioThink-SFT} &\underline{74.45} & 72.13 &\underline{73.28} &	\underline{66.18} &	\underline{62.16} &	\underline{64.11} &	48.32 &	41.20 &\underline{44.47} &\underline{89.85} &\underline{88.78} &\underline{89.31} &\underline{67.17} &64.22 &\underline{65.66} &\underline{67.15} &	\textbf{60.12} &\underline{63.44} \\
\textbf{CardioThink-SSPO} &\textbf{75.46} &\underline{72.92} &\textbf{74.17} &\textbf{69.50} &\textbf{64.73} &\textbf{67.03} &\underline{55.69} &\textbf{45.75} &\textbf{50.23} &\textbf{90.67} &\textbf{90.11} &\textbf{90.39}  &\textbf{68.06} &\textbf{65.87} &\textbf{66.95} &\textbf{74.00} &\underline{56.64} &	\textbf{64.16}  \\
\hline
\textbf{Gain} & +7.25 &	-2.38 &	+4.28 &	+9.34 &	+14.57 &	+20.18 &	-8.75 &	+0.68 &	+17.32 &	+7.38 &	+6.82 &	+7.1 &	+2.2 &	+0.83 &	+3.09 &	+30.63 &	+13.24 &	+22.73\\
\bottomrule
\end{tabular}
} 
\label{mainresults}
\vspace{-0.5cm}
\end{table*}

\section{Results and Discussion}

\subsection{Comparison with State-of-the-Art Methods}
Table~\ref{mainresults} compares CardioThink with a broad range of state-of-the-art ECG classification methods across six benchmark tasks, including pretrained discriminative models and recent multimodal large language model–based approaches that jointly leverage ECG signals, images, and clinical text.

Across all tasks, CardioThink consistently achieves the highest F1 scores, outperforming all competing methods with an average F1 improvement of 12.45\%. This improvement is particularly important in clinical ECG diagnosis, where F1 better reflects diagnostic reliability under class imbalance.

\noindent\textbf{Comparison with Pretrained Discriminative Models.}
Compared with pretrained discriminative methods such as MERL and MELP, which rely on representation learning followed by direct label prediction, CardioThink achieves consistent improvements across all PTB-XL subtasks as well as CPSC2018 and CSN. This suggests that relying solely on direct label prediction from learned representations may not fully capture the complexity of clinical decision-making, and that explicitly modeling diagnostic reasoning provides complementary benefits beyond pretraining.

\noindent\textbf{Comparison with Generative Models.}
CardioThink also outperforms generative and multimodal large language model–based methods, including QoQ-Med, GEM, PULSE, and ECG-Chat, yielding particularly large gains on fine-grained tasks such as PTBXL-Sub and PTBXL-Form. Although existing models leverage multimodal alignment and language modeling, their reasoning toward a diagnosis remains implicit and unstructured. By explicitly decomposing ECG interpretation into rhythm, conduction, morphology, and impression, CardioThink introduces a structured reasoning process that proves to be a more efficient and accurate approach to automated diagnosis.

\noindent\textbf{Effect of Structured Set Policy Optimization.}
Comparing CardioThink-SFT and CardioThink-SSPO demonstrates the effectiveness of Structured Set Policy Optimization. 
Leveraging reinforcement learning, SSPO explicitly optimizes reasoning quality by simultaneously enhancing reasoning structure and diagnostic accuracy. In contrast to SFT, which is confined to imitation, SSPO utilizes reward signals to enforce strict structural constraints and refine logical chains. Consequently, this mechanism yields superior performance over the supervised baseline.

\begin{wraptable}{r}{0.5\textwidth} 
\vspace{-15pt} 
\caption{Ablation study on the PTB-XL dataset (Subclass).}
\label{tab:ablation_study_new}
\centering
\small
\setlength{\tabcolsep}{3pt} 
\scalebox{0.85}{ 
\begin{tabular}{llccc}
\toprule
\multirow{2}{*}{Phase} & \multirow{2}{*}{Method} & \multicolumn{3}{c}{ptbxl-sub} \\
\cmidrule(lr){3-5}
& & Precision & Recall & F1-score \\
\midrule
\multirow{3}{*}{Cold Start} 
& w/o Think &63.93 &62.56 &63.24 \\
& Original Think &64.51 &62.20 &63.33 \\
& Structure Think &66.18 &62.16 &64.11 \\
\midrule
\multirow{3}{*}{SSPO} 
& w/o Structure &66.02 &61.60 &63.73 \\
& w/o Diagnosis &69.20 &60.81 &64.74 \\
& \textbf{Full} &\textbf{69.50} &\textbf{64.73} &\textbf{67.03} \\
\bottomrule
\end{tabular}
} 
\vspace{-10pt} 
\end{wraptable}

\begin{figure*}[t!]
    \centering
    \includegraphics[width=0.9\linewidth]{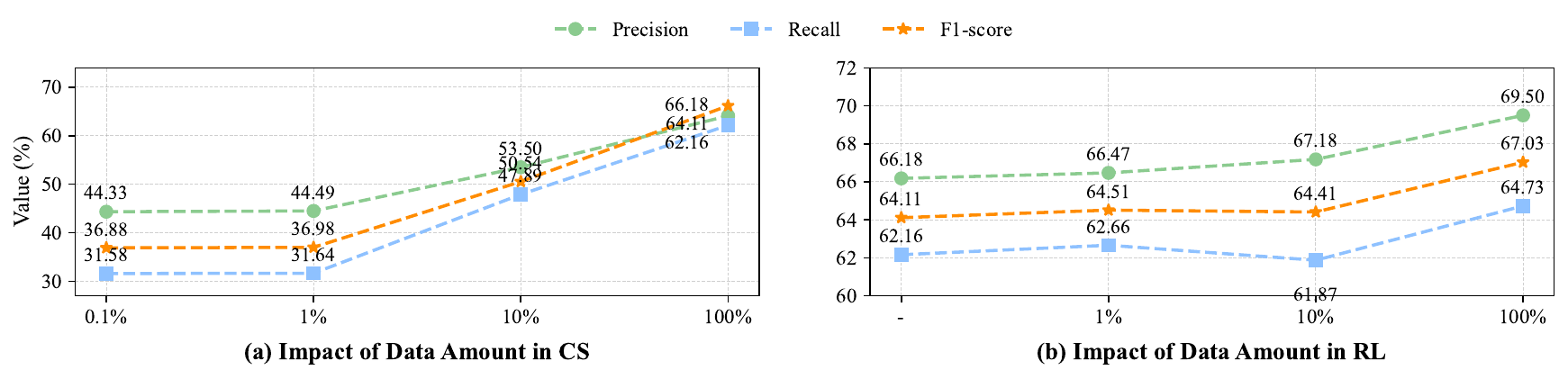}
    \caption{Impact of training data amount on performance under (a) supervised fine-tuning (CS) and (b) reinforcement learning with SSPO (RL), evaluated on the PTB-XL Subclass task.}
    \label{fig:data amoun exp}
    \vspace{-0.5cm}
\end{figure*}

\subsection{Ablation Study}

Table~\ref{tab:ablation_study_new} evaluates the impact of various cold-start thinking strategies and distinct reward components within SSPO on the PTB-XL Subclass.

\textbf{Effect of Reasoning Structure.} Under the {Cold Start} setting, introducing reasoning consistently improves performance over direct prediction (\emph{w/o Think}). While free-form reasoning (\emph{Original Think}) yields moderate gains, enforcing a clinically structured reasoning process (\emph{Structure Think}) achieves the highest F1-score, demonstrating that reasoning effectiveness depends on explicit structural constraints rather than reasoning generation alone.

\textbf{Effect of Reward Components in SSPO.} Applying SSPO further boosts performance. Removing either the reasoning format constraint (\emph{w/o Structure}) or the set-level accuracy objective (\emph{w/o Diagnosis}) leads to clear degradation, indicating that both components contribute substantially. The full SSPO configuration delivers the best Precision, Recall, and F1-score, confirming that jointly optimizing structured reasoning adherence and variable-size diagnostic accuracy is critical for effective learning.

Overall, these results show that explicitly modeling and optimizing clinical reasoning is essential not only for interpretability, but also for achieving superior ECG classification performance.

\textbf{Impact of Training Data Amount.}
We analyze the effect of training data scale during the Cold Start (CS) and reinforcement learning (RL) stages. As shown in Figure.~\ref{fig:data amoun exp} (a), CardioThink performs poorly in low-data regimes ($0.1\%$ and $1\%$) with F1-scores below 37\%, highlighting the data dependency of supervised learning. Performance improves substantially, reaching 64.11\% on the full dataset. During the RL stage, Figure.~\ref{fig:data amoun exp} (b) demonstrates that SSPO consistently improves performance across all data scales. Even with limited RL data, SSPO outperforms the CS baseline, confirming the effectiveness of reward-based optimization. The best overall performance (F1 = 67.03\%) is achieved using the full RL data.
Overall, while cold start mainly benefits from increased data volume, SSPO provides complementary and data-efficient gains by explicitly optimizing structured clinical reasoning.

\subsection{Reasoning Quality Evaluation}
\label{qualitys}
We employ a dual-evaluation approach to comprehensively assess the quality of clinical reasoning generated by CardioThink on the \textit{PTB-XL} Subclass test set. First, following prior work~\cite{GEM}, we prompt GPT-4o~\cite{hurst2024gpt} to act as an expert cardiologist and evaluate the generated reasoning across the entire test set. Second, to validate this LLM-as-a-Judge framework, three medical experts were compensated to evaluate a random 10\% sample of the test set, specifically assessing the reasoning generated by the SSPO model. Since existing generative ECG models do not produce reasoning traces that directly support diagnostic classification, we compare CardioThink's Cold Start (SFT) and SSPO to isolate the impact of reinforcement learning.

We propose five metrics to evaluate reasoning quality: (1) Syntactic Structural Validity (SSV) ensures strict adherence to the predefined format; (2) Ground-Truth Feature Alignment (GTFA) measures semantic alignment by verifying the anatomical localization of pathological features; (3) Semantic Disentanglement (SD) assesses informational orthogonality to prevent cross-domain contamination; (4) Deductive Logic Consistency (DLC) evaluates internal coherence, ensuring observations causally necessitate the diagnosis; and (5) Evidence Specificity (ES) gauges the granularity of clinical data, prioritizing quantitative precision over templated descriptions. While the LLM evaluates all five metrics, the human assessment focuses exclusively on the three clinically pivotal ones (GTFA, DLC, and ES), excluding SSV and SD as they primarily assess structural and linguistic characteristics robustly handled by the LLM.

As shown in Table~\ref{tab:llm_judge_scores}, CardioThink-SSPO consistently outperforms the SFT baseline. While the scores of the SFT model already reflect a baseline of clinical soundness, the further gains achieved by SSPO confirm its ability to refine the clinical validity of the generated rationales. This demonstrates that SSPO goes beyond merely boosting predictive accuracy to fundamentally enhance the underlying diagnostic reasoning. Notably, the most critical improvement is observed in {GTFA}, indicating that SSPO significantly strengthens the model's ability to accurately localize pathological evidence within the correct clinical domains. Furthermore, SSPO achieves perfect compliance in {SSV}, effectively enforcing strict adherence to the predefined reasoning schema, which is difficult to guarantee through SFT alone. Improvements in {SD} reduce cross-component leakage, while noticeable gains in {DLC} and {ES} confirm that the generated observations form a coherent, causally sound diagnostic process utilizing clinically precise data rather than surface-level pattern matching.

As reported in Table~\ref{tab:llm_judge_scores_right}, the close alignment between human and LLM mean scores across the three evaluated metrics further demonstrates that the generated reasoning has indeed achieved a certain degree of clinical validity. Furthermore, the human experts exhibited substantial inter-rater reliability, achieving a Fleiss' $\kappa$ between $0.617$ and $0.643$. Crucially, we observe strong relative alignment between the expert evaluations and the LLM judge, evidenced by high Quadratic Weighted Kappa (QWK) and Spearman correlation coefficients. Together, these robust consistency metrics substantiate that our automated evaluation reliably proxies professional medical judgment.

\subsection{Case Study}

\begin{figure*}[t!]
    \centering
    \includegraphics[width=1\linewidth]{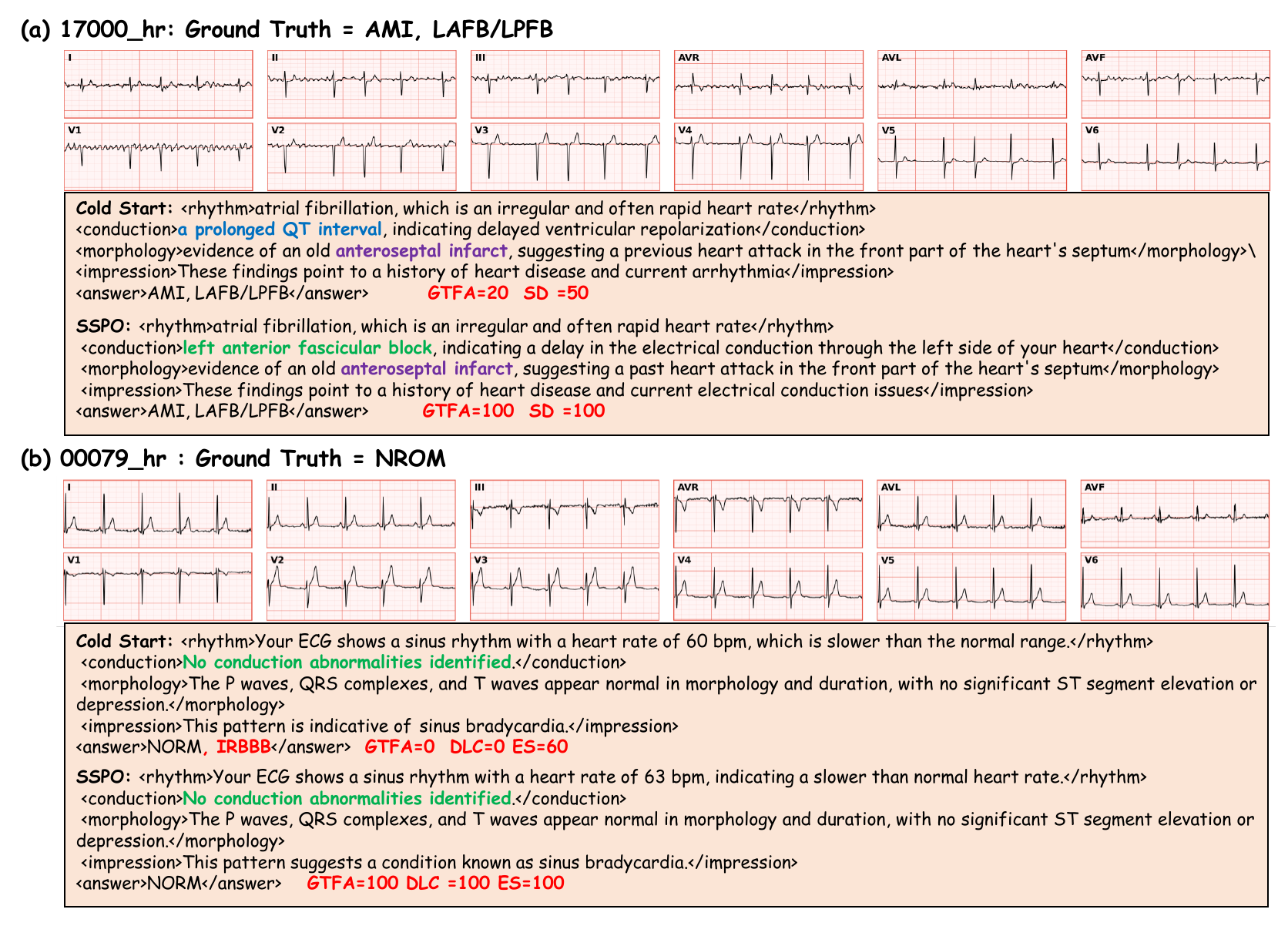}
    \caption{Comparison of reasoning quality between the Cold-Start model and the SSPO-aligned model. (a) Abnormal ECG sample with pathological findings. (b) Normal ECG sample. The results highlight the corrections and logical refinements introduced by SSPO.}
    \label{fig:case study}
    \vspace{-0.5cm}
\end{figure*}

Figure~\ref{fig:case study} illustrates SSPO's impact on diagnostic accuracy and interpretability using representative cases from the PTB-XL Subclass test set. These cases demonstrate that by optimizing reasoning via reinforcement learning, CardioThink-SSPO mitigates the spurious predictions inherent in the cold-start stage, ensuring robust ECG classification. Detailed case analyses are provided below:

\textbf{Case 1: Superior Reasoning for Correct Prediction.}  
For sample \textit{17000\_hr} (Ground Truth: \textit{AMI, LAFB/LPFB}), both models predict the correct labels, but their reasoning quality differs significantly. The Cold-Start model misattributes a prolonged QT interval and fails to identify the underlying conduction abnormality (GTFA = 20). Conversely, CardioThink-SSPO explicitly recognizes the \textit{left anterior fascicular block} and consistently associates infarction evidence with morphological changes. This results in perfect GTFA and semantic disentanglement, demonstrating a reasoning process that is more clinically grounded and better aligned with expert interpretation.

\textbf{Case 2: Improved Reasoning Correcting Diagnosis.}  
For sample \textit{00079\_hr} (Ground Truth: \textit{NORM}), the Cold-Start model generates generic reasoning and incorrectly diagnoses \textit{IRBBB}, leading to poor feature alignment (GTFA = 0) and inconsistent logic (DLC = 0). In contrast, CardioThink-SSPO provides precise and restrained analysis, correctly identifying sinus bradycardia as a non-pathological variation. By excluding spurious conduction abnormalities, SSPO achieves the correct diagnosis with perfect GTFA and DLC scores.

\begin{table}[t!]
    \centering
    \newsavebox{\righttablebox}
    \sbox{\righttablebox}{%
        \setlength{\tabcolsep}{3pt}
        \begin{tabular}{lcccc}
            \toprule
            \textbf{Metric} & \textbf{Mean} &\textbf{Fleiss' $\kappa$} & \textbf{QWK} & \textbf{Spearman} \\
            \midrule
            Ground-Truth Feat. Align.  & 62.90& 0.639 & 0.793 & 0.736 \\
            Deductive Logic Consistency & 70.25& 0.617 & 0.789 & 0.729 \\
            Evidence Specificity        & 68.30& 0.643 & 0.734 & 0.751 \\
            \bottomrule
        \end{tabular}%
    }

    \begin{minipage}[t]{0.48\textwidth}
        \centering
        \caption{LLM-as-a-Judge evaluation on the PTB-XL dataset (\%).}
        \label{tab:llm_judge_scores}
        \vspace{2mm}
        
        \setlength{\tabcolsep}{3pt}
        \resizebox{\dimexpr \width * \linewidth / \wd\righttablebox \relax}{!}{%
            \begin{tabular}[t]{lcc}
                \toprule
                \textbf{Metric} & \textbf{SFT} & \textbf{SSPO} \\
                \midrule
                Syntactic Struct. Val.      & 98.10  & \textbf{100.00} \\
                Ground-Truth Feat. Align.   & 60.97 & \textbf{63.18}  \\
                Semantic Disentanglement    & 86.10 & \textbf{88.92}  \\
                Deductive Logic Consistency & 67.66 & \textbf{69.80}  \\
                Evidence Specificity        & 67.00 & \textbf{68.38}  \\
                \bottomrule
            \end{tabular}%
        }
    \end{minipage}%
    \hfill 
    \begin{minipage}[t]{0.48\textwidth}
        \centering
        \caption{Human metrics on PTB-XL dataset of SSPO reasonings.}
        \label{tab:llm_judge_scores_right}
        \vspace{2mm}
        
        \resizebox{\linewidth}{!}{%
            \usebox{\righttablebox}
        }
    \end{minipage}
    \vspace{-0.5cm}
\end{table}

\section{Conclusion}
This work introduces CardioThink, a novel framework that bridges the gap between black-box predictions and clinical logic by explicitly modeling structured reasoning within MLLMs. We developed a reasoning-oriented ECG dataset with structured diagnostic components and introduced SSPO to synergistically optimize reasoning quality and diagnostic accuracy. Empirical results on six large-scale benchmarks reveal that CardioThink surpasses existing multimodal models by an average of 12.45\% in F1-score. Notably, reasoning quality evaluations confirm that while structured reasoning provides a solid foundation of clinical soundness, SSPO further elevates the clinical validity and logical precision of the generated rationales. Critically, our analysis establishes a direct link between reasoning fidelity and prediction reliability. Ultimately, CardioThink illustrates that integrating structured clinical logic offers a valuable approach for generative ECG analysis, paving the way for more reliable and interpretable clinical decision support systems.

\bibliographystyle{plainnat}
\bibliography{custom}

@article{yang2025qwen3,
  title={Qwen3 technical report},
  author={Yang, An and Li, Anfeng and Yang, Baosong and Zhang, Beichen and Hui, Binyuan and Zheng, Bo and Yu, Bowen and Gao, Chang and Huang, Chengen and Lv, Chenxu and others},
  journal={arXiv preprint arXiv:2505.09388},
  year={2025}
}

@article {Brushe064389,
	author = {Brush, John E and Sherbino, Jonathan and Norman, Geoffrey R},
	title = {Diagnostic reasoning in cardiovascular medicine},
	volume = {376},
	elocation-id = {e064389},
	year = {2022},
	doi = {10.1136/bmj-2021-064389},
	publisher = {BMJ Publishing Group Ltd},
	URL = {https://www.bmj.com/content/376/bmj-2021-064389},
	eprint = {https://www.bmj.com/content/376/bmj-2021-064389.full.pdf},
	journal = {BMJ}
}

@article{hurst2024gpt,
  title={Gpt-4o system card},
  author={Hurst, Aaron and Lerer, Adam and Goucher, Adam P and Perelman, Adam and Ramesh, Aditya and Clark, Aidan and Ostrow, AJ and Welihinda, Akila and Hayes, Alan and Radford, Alec and others},
  journal={arXiv preprint arXiv:2410.21276},
  year={2024}
}

@article{wagner2020ptb,
  title={PTB-XL, a large publicly available electrocardiography dataset},
  author={Wagner, Patrick and Strodthoff, Nils and Bousseljot, Ralf-Dieter and Kreiseler, Dieter and Lunze, Fatima I and Samek, Wojciech and Schaeffter, Tobias},
  journal={Scientific data},
  volume={7},
  number={1},
  pages={1--15},
  year={2020},
  publisher={Nature Publishing Group}
}

@article{liu2018open,
  title={An open access database for evaluating the algorithms of electrocardiogram rhythm and morphology abnormality detection},
  author={Liu, Feifei and Liu, Chengyu and Zhao, Lina and Zhang, Xiangyu and Wu, Xiaoling and Xu, Xiaoyan and Liu, Yulin and Ma, Caiyun and Wei, Shoushui and He, Zhiqiang and others},
  journal={Journal of Medical Imaging and Health Informatics},
  volume={8},
  number={7},
  pages={1368--1373},
  year={2018},
  publisher={American Scientific Publishers}
}

@article{zheng2020optimal,
  title={Optimal multi-stage arrhythmia classification approach},
  author={Zheng, Jianwei and Chu, Huimin and Struppa, Daniele and Zhang, Jianming and Yacoub, Sir Magdi and El-Askary, Hesham and Chang, Anthony and Ehwerhemuepha, Louis and Abudayyeh, Islam and Barrett, Alexander and others},
  journal={Scientific reports},
  volume={10},
  number={1},
  pages={2898},
  year={2020},
  publisher={Nature Publishing Group UK London}
}

@article{zheng2022large,
  title={A large scale 12-lead electrocardiogram database for arrhythmia study (version 1.0. 0)},
  author={Zheng, Jianwei and Guo, Hangyuan and Chu, Huimin},
  journal={PhysioNet 2022Available online httpphysionet orgcontentecg arrhythmia10 0accessed on},
  volume={23},
  pages={7},
  year={2022}
}

@article{hendrycks2016gaussian,
  title={Gaussian Error Linear Units (Gelus)},
  author={Hendrycks, D},
  journal={arXiv preprint arXiv:1606.08415},
  year={2016}
}

@inproceedings{liu2024improved,
  title={Improved baselines with visual instruction tuning},
  author={Liu, Haotian and Li, Chunyuan and Li, Yuheng and Lee, Yong Jae},
  booktitle={Proceedings of the IEEE/CVF conference on computer vision and pattern recognition},
  pages={26296--26306},
  year={2024}
}

@inproceedings{ECGCHAT,
  title={Ecg-chat: A large ecg-language model for cardiac disease diagnosis},
  author={Zhao, Yubao and Kang, Jiaju and Zhang, Tian and Han, Puyu and Chen, Tong},
  booktitle={2025 IEEE International Conference on Multimedia and Expo (ICME)},
  pages={1--6},
  year={2025},
  organization={IEEE}
}

@article{yuan2025enhancing,
  title={Enhancing Multi-Label ECG Classification via Task-Guided Lead Correlations in Internet of Medical Things},
  author={Yuan, Xiaoyan and Wang, Wei and Chen, Junxin and Fang, Kai and Bashir, Ali Kashif and Mondal, Tapas and Hu, Xiping and Deen, M Jamal},
  journal={IEEE Internet of Things Journal},
  year={2025},
  publisher={IEEE}
}

@inproceedings{yuan2025reading,
  title={Reading Between the Channels: Knowledge-Augmented Medical Time Series Classification},
  author={Yuan, Xiaoyan and Wang, Wei and Chen, Junxin and Hu, Xiping},
  booktitle={Proceedings of the 33rd ACM International Conference on Multimedia},
  pages={8978--8987},
  year={2025}
}

@inproceedings{ECG2TOK,
  title={ECG2TOK: ECG Pre-Training with Self-Distillation Semantic Tokenizers},
  author={Yuan, Xiaoyan and Wang, Wei and Liu, Han and Chen, Jian and Hu, Xiping},
  booktitle={34th Internationa Joint Conference on Artificial Intelligence, IJCAI 2025},
  pages={9990--9998},
  year={2025},
  organization={International Joint Conferences on Artificial Intelligence}
}

@inproceedings{HeartLang,
  title={Reading Your Heart: Learning ECG Words and Sentences via Pre-training ECG Language Model},
  author={Jin, Jiarui and Wang, Haoyu and Li, Hongyan and Li, Jun and Pan, Jiahui and Hong, Shenda},
  booktitle={The Thirteenth International Conference on Learning Representations}
}

@article{tracing,
  title={Tracing the Heart's Pathways: ECG Representation Learning from a Cardiac Conduction Perspective},
  author={Pan, Tan and Sun, Yixuan and Jiang, Chen and Gao, Qiong and Sun, Rui and Zhang, Xingmeng and Yang, Zhenqi and Han, Limei and Liang, Yixiu and Cheng, Yuan and others},
  journal={arXiv preprint arXiv:2512.24002},
  year={2025}
}

@inproceedings{MERL,
  title={Zero-Shot ECG Classification with Multimodal Learning and Test-time Clinical Knowledge Enhancement},
  author={Liu, Che and Wan, Zhongwei and Ouyang, Cheng and Shah, Anand and Bai, Wenjia and Arcucci, Rossella},
  booktitle={Forty-first International Conference on Machine Learning}
}

@article{SuPreME,
  title={SuPreME: A Supervised Pre-training Framework for Multimodal ECG Representation Learning},
  author={Cai, Mingsheng and Jiang, Jiuming and Huang, Wenhao and Liu, Che and Arcucci, Rossella},
  journal={arXiv preprint arXiv:2502.19668},
  year={2025}
}

@article{GEM,
  title={Gem: Empowering mllm for grounded ecg understanding with time series and images},
  author={Lan, Xiang and Wu, Feng and He, Kai and Zhao, Qinghao and Hong, Shenda and Feng, Mengling},
  journal={arXiv preprint arXiv:2503.06073},
  year={2025}
}

@article{liu2025fleming,
  title={Fleming-r1: Toward expert-level medical reasoning via reinforcement learning},
  author={Liu, Chi and Li, Derek and Shu, Yan and Chen, Robin and Duan, Derek and Fang, Teng and Dai, Bryan},
  journal={arXiv preprint arXiv:2509.15279},
  year={2025}
}

@article{yang2025ecg,
  title={ECG-LM: Understanding Electrocardiogram with a Large Language Model},
  author={Yang, Kai and Hong, Massimo and Zhang, Jiahuan and Luo, Yizhen and Zhao, Suyuan and Zhang, Ou and Yu, Xiaomao and Zhou, Jiawen and Yang, Liuqing and Zhang, Ping and others},
  journal={Health Data Science},
  volume={5},
  pages={0221},
  year={2025},
  publisher={AAAS}
}

@article{jin2025uniecg,
  title={Uniecg: Understanding and generating ecg in one unified model},
  author={Jin, Jiarui and Wang, Haoyu and Lan, Xiang and Li, Jun and Cheng, Gaofeng and Li, Hongyan and Hong, Shenda},
  journal={arXiv preprint arXiv:2509.18588},
  year={2025}
}

@article{schulman2017proximal,
  title={Proximal policy optimization algorithms},
  author={Schulman, John and Wolski, Filip and Dhariwal, Prafulla and Radford, Alec and Klimov, Oleg},
  journal={arXiv preprint arXiv:1707.06347},
  year={2017}
}

@article{HE2025102963,
title = {A survey of large language models for healthcare: from data, technology, and applications to accountability and ethics},
journal = {Information Fusion},
volume = {118},
pages = {102963},
year = {2025},
issn = {1566-2535},
doi = {https://doi.org/10.1016/j.inffus.2025.102963},
url = {https://www.sciencedirect.com/science/article/pii/S1566253525000363},
author = {Kai He and Rui Mao and Qika Lin and Yucheng Ruan and Xiang Lan and Mengling Feng and Erik Cambria}
}

@article{li2025generative,
  title={Generative classifiers avoid shortcut solutions},
  author={Li, Alexander C and Kumar, Ananya and Pathak, Deepak},
  journal={arXiv preprint arXiv:2512.25034},
  year={2025}
}

@inproceedings{zhou2025robustness,
  title={Robustness to spurious correlations via dynamic knowledge transfer},
  author={Zhou, Xiaoling and Ye, Wei and Lee, Zhemg and Zhang, Shikun},
  booktitle={Proceedings of the Thirty-Fourth International Joint Conference on Artificial Intelligence},
  pages={7182--7190},
  year={2025}
}

@article{jin2026ecg,
  title={ECG-R1: Protocol-Guided and Modality-Agnostic MLLM for Reliable ECG Interpretation},
  author={Jin, Jiarui and Wang, Haoyu and Wu, Xingliang and Fang, Xiaocheng and Lan, Xiang and Wang, Zihan and Zhang, Deyun and Liu, Bo and Zhang, Yingying and Wu, Xian and others},
  journal={arXiv preprint arXiv:2602.04279},
  year={2026}
}

@article{yu2025dapo,
  title={Dapo: An open-source llm reinforcement learning system at scale},
  author={Yu, Qiying and Zhang, Zheng and Zhu, Ruofei and Yuan, Yufeng and Zuo, Xiaochen and Yue, Yu and Dai, Weinan and Fan, Tiantian and Liu, Gaohong and Liu, Lingjun and others},
  journal={arXiv preprint arXiv:2503.14476},
  year={2025}
}

@inproceedings{
dai2026qoqmed,
title={QoQ-Med: Building Multimodal Clinical Foundation Models with Domain-Aware {GRPO} Training},
author={Wei Dai and Peilin Chen and Chanakya Ekbote and Paul Pu Liang},
booktitle={The Thirty-ninth Annual Conference on Neural Information Processing Systems},
year={2026},
url={https://openreview.net/forum?id=ZwCVFBFUFb}
}

@inproceedings{DBEAT,
  title={Boosting Masked ECG-Text Auto-Encoders as Discriminative Learners},
  author={Hung, Manh Pham and Saeed, Aaqib and Ma, Dong},
  booktitle={Forty-second International Conference on Machine Learning}
}

@article{shao2024deepseekmath,
  title={Deepseekmath: Pushing the limits of mathematical reasoning in open language models},
  author={Shao, Zhihong and Wang, Peiyi and Zhu, Qihao and Xu, Runxin and Song, Junxiao and Bi, Xiao and Zhang, Haowei and Zhang, Mingchuan and Li, YK and Wu, Yang and others},
  journal={arXiv preprint arXiv:2402.03300},
  year={2024}
}

@inproceedings{
hu2022lora,
title={Lo{RA}: Low-Rank Adaptation of Large Language Models},
author={Edward J Hu and yelong shen and Phillip Wallis and Zeyuan Allen-Zhu and Yuanzhi Li and Shean Wang and Lu Wang and Weizhu Chen},
booktitle={International Conference on Learning Representations},
year={2022},
url={https://openreview.net/forum?id=nZeVKeeFYf9}
}

@inproceedings{MELP,
  title={From Token to Rhythm: A Multi-Scale Approach for ECG-Language Pretraining},
  author={Wang, Fuying and Xu, Jiacheng and Yu, Lequan},
  booktitle={Forty-second International Conference on Machine Learning}
}

@inproceedings{MEIT,
  title={MEIT: Multimodal Electrocardiogram Instruction Tuning on Large Language Models for Report Generation},
  author={Wan, Zhongwei and Liu, Che and Wang, Xin and Tao, Chaofan and Shen, Hui and Xiong, Jing and Arcucci, Rossella and Yao, Huaxiu and Zhang, Mi},
  booktitle={Findings of the Association for Computational Linguistics: ACL 2025},
  pages={14510--14527},
  year={2025}
}

@article{QHEART,
  title={Q-Heart: ECG Question Answering via Knowledge-Informed Multimodal LLMs},
  author={Pham, Hung Manh and Tang, Jialu and Saeed, Aaqib and Ma, Dong},
  journal={arXiv preprint arXiv:2505.06296},
  year={2025}
}

@article{PULS,
  title={Teach multimodal llms to comprehend electrocardiographic images},
  author={Liu, Ruoqi and Bai, Yuelin and Yue, Xiang and Zhang, Ping},
  journal={arXiv preprint arXiv:2410.19008},
  year={2024}
}

@article{MVMNet,
  title={A multi-view multi-scale neural network for multi-label ECG classification},
  author={Yang, Shunxiang and Lian, Cheng and Zeng, Zhigang and Xu, Bingrong and Zang, Junbin and Zhang, Zhidong},
  journal={IEEE Transactions on Emerging Topics in Computational Intelligence},
  volume={7},
  number={3},
  pages={648--660},
  year={2023},
  publisher={IEEE}
}

@inproceedings{MRMNet,
  title={A multi-resolution mutual learning network for multi-label ecg classification},
  author={Huang, Wei and Wang, Ning and Feng, Panpan and Wang, Haiyan and Wang, Zongmin and Zhou, Bing},
  booktitle={2024 IEEE International Conference on Bioinformatics and Biomedicine (BIBM)},
  pages={3303--3306},
  year={2024},
  organization={IEEE}
}

@article{naturereview,
  title={Artificial intelligence-enhanced electrocardiography in cardiovascular disease management},
  author={Siontis, Konstantinos C and Noseworthy, Peter A and Attia, Zachi I and Friedman, Paul A},
  journal={Nature Reviews Cardiology},
  volume={18},
  number={7},
  pages={465--478},
  year={2021},
  publisher={Nature Publishing Group UK London}
}

@article{NC,
  title={Automatic diagnosis of the 12-lead ECG using a deep neural network},
  author={Ribeiro, Ant{\^o}nio H and Ribeiro, Manoel Horta and Paix{\~a}o, Gabriela MM and Oliveira, Derick M and Gomes, Paulo R and Canazart, J{\'e}ssica A and Ferreira, Milton PS and Andersson, Carl R and Macfarlane, Peter W and Meira Jr, Wagner and others},
  journal={Nature communications},
  volume={11},
  number={1},
  pages={1760},
  year={2020},
  publisher={Nature Publishing Group UK London}
}

@article{naturemedicine,
  title={Artificial intelligence for direct-to-physician reporting of ambulatory electrocardiography},
  author={Johnson, LS and Zadrozniak, P and Jasina, G and Grotek-Cuprjak, A and Andrade, JG and Svennberg, E and Diederichsen, SZ and McIntyre, WF and Stavrakis, S and Benezet-Mazuecos, J and others},
  journal={Nature Medicine},
  volume={31},
  number={3},
  pages={925--931},
  year={2025},
  publisher={Nature Publishing Group US New York}
}

\end{document}